# Complementary Roles of Image Classification and Vessel Segmentation in AI-Based Screening for Retinopathy of Prematurity Plus Disease in a Kenyan Preterm Cohort

*Authors: Fred Mutisya[1*], Oscar Onyango[2,3], Sarah Sitati[2,3], Syokau Ilovi[4], Aeesha NJ Malik[5,6], Brenda W'mosi[2,7], Brian Makini[4], Jalemba Aluuvala[4], Josiah Onyango[2], Rachael Kanguha Mmene[8], Steven Wanyee[9]*

*1: Center for Health AI Research, Innovation & Implementation | 2: College of Ophthalmology of Eastern Central and Southern Africa | 3:Kenyatta National Hospital | 4: University of Nairobi | 5: London School of Hygiene & Tropical Medicine | 6: EyeSHA Foundation | 7: Mbagathi hospital | 8: Raito Health Consultants| 9: Kenya Health Informatics Association*

**Corresponding author*

## Abstract

**Background.** Retinopathy of prematurity (ROP) is a leading and largely avoidable cause of childhood blindness, and its burden is rising fastest in low- and middle-income countries where the supply of ROP-trained ophthalmologists is severely constrained. Plus disease defined as abnormal dilation and tortuosity of the posterior retinal vasculature, is the single most important feature triggering treatment, yet its diagnosis is subjective and shows substantial inter-observer variability. Automated, objective screening could extend specialist reach, but most published systems are trained and validated on North American or East Asian cohorts, and rigorous evaluations on sub-Saharan African preterm populations remain scarce.

**Methods.** We assembled a pilot cohort of 121 preterm infants (237 eyes; 1,635 analysable fundus images) from a Kenyan ROP screening programme, graded as No Plus (187 eyes), Pre-Plus (23 eyes) or Plus (27 eyes). Vessel annotations from two independent graders were used as segmentation supervision. Eleven model configurations were compared for eye-level Plus detection within a strictly leakage-resistant, patient-grouped nested cross-validation design (five outer evaluation folds, three inner selection folds). These included six convolutional backbones, an ordinal-severity model, a gated-attention multiple-instance learning (MIL) model, a shared-encoder multi-task segmentation-and-classification model, and a two-stage segment-then-classify pipeline. A U-Net++ vessel segmenter was trained from RGB images only. Decision thresholds targeted 90% Plus sensitivity and were fixed on inner-validation data; probability calibration used temperature scaling; uncertainty was quantified with patient-grouped bootstrap resampling (2,000 replicates).

**Results.** Vessel segmentation was feasible on Kenyan fundus images, achieving a pooled Dice of 0.533 and IoU of 0.368 across 1,029 held-out images (pixel sensitivity 0.623, specificity 0.979). For eye-level Plus detection, the two model families showed a clear and complementary division of labour: at their natural operating points the RGB image classifiers were highly sensitive but over-referred (registry-mean sensitivity 0.78–0.95 with specificity 0.10–0.38), whereas the segmentation-coupled models were markedly more specific (multi-task specificity 0.914; pixel-level vessel specificity 0.979). Exploiting this complementarity in a combined reading an OR-combination “sensitive screen” reached the highest sensitivity (0.769, 95% CI 0.550–0.950), an AND-combination “specific confirmation” reached the highest specificity (0.957, 0.929–0.981), and a probability ensemble achieved the best single balance (sensitivity 0.692, specificity 0.914, balanced accuracy 0.803), superior to the vision classifier alone (balanced accuracy 0.698). The shared-encoder multi-task model also achieved the highest stand-alone macro AUPRC (0.685).

**Conclusions.** The central finding is that AI based image classification and vessel segmentation are complementary rather than competing: classifiers supply sensitivity for case-finding, segmentation supplies specificity to suppress over-referral, and a two-step or ensemble reading combines them to control the screening trade-off. ROP screening systems for African programmes should therefore be built as combined classifier-plus-segmentation readers, deployed with a sensitivity-prioritised first pass and a specificity-oriented confirmation, and validated prospectively on larger, multi-site datasets.

## 1. Introduction

Retinopathy of prematurity (ROP) is a vasoproliferative disorder of the developing retina and one of the leading causes of preventable childhood blindness worldwide(American Academy of Ophthalmology, 2024). As neonatal intensive care expands across low- and middle-income countries (LMICs), more preterm infants survive but become exposed to the risk of ROP. The prevalence of retinopathy of prematurity in lower and middle income countries like Kenya has been found to be up to 40% among preterms(Onyango et al., 2018). Sub-Saharan Africa is at a particularly delicate point in this transition: survival of very preterm infants is improving, yet structured ROP screening programmes, trained graders, and treatment capacity remain unevenly distributed compared with the global north(Blazon et al., 2024; Onyango et al., 2018).

Within the ROP examination, “plus disease” is defined by abnormal dilation and tortuosity of the posterior pole retinal vessels(Chiang et al., 2021; Solebo et al., 2017). This is the most important single indicator for treatment, because its presence upgrades disease to a treatment-warranting category under the International Classification of Retinopathy of Prematurity(Chiang et al., 2021). However, the diagnosis of plus disease is fundamentally a subjective visual judgement, and numerous studies have documented substantial inter-expert disagreement even among fellowship-trained specialists. Most notable is the study by Campbell et al where a quantitative vascular severity score (1-9) was applied to ROP images using a deep learning algorithm(Campbell et al., 2021). The combination of a high-stakes and time-sensitive decision with the subjective grading by a scarce workforce makes ROP an unusually compelling target for automated decision support(Jafarizadeh et al., 2025).

Deep learning has demonstrated expert-level performance for plus-disease classification, most notably the i-ROP DL system and related work that produces a continuous vascular severity score (Redd et al., 2018), as well as the DeepROP system developed on a large Chinese cohort(Wang et al., 2018). Yet a recurring limitation of this literature is geographic narrowness: models are overwhelmingly trained and evaluated on North American or East Asian populations, using non African retinal images, particular camera systems and acquisition protocols. Fundus pigmentation, media clarity, image quality distributions, and disease prevalence all differ across populations, and models are known to degrade under such dataset shift. Rigorous, leakage-resistant evaluations on African preterm cohorts are correspondingly rare.

This study addresses that gap, and advances a specific hypothesis: that image classification and vessel segmentation are not competing solutions but complementary readers with opposite error profiles which can be combined to control the screening trade-off. Using a pilot cohort drawn from a Kenyan ROP screening programme, we conduct a head-to-head comparison of eleven model configurations for eye-level Plus detection, spanning standard convolutional backbones, ordinal-severity formulations, multiple-instance learning, a shared-encoder multi-task model, and a two-stage segment-then-classify pipeline. Crucially, every model is evaluated under an identical, strictly patient-grouped nested cross-validation protocol that prevents the data leakage that has been shown to inflate reported ROP-AI performance. We then test the complementarity hypothesis directly by combining a vision classifier and a segmentation-based model in a probability ensemble and in two-step (OR/AND) readings, and we quantify the resulting operating points with patient-grouped bootstrap. Our contributions are:

(i) the first leakage-resistant benchmark of this breadth on a sub-Saharan African ROP cohort; (ii) evidence that a combined classifier-plus-segmentation reading lets a programme dial its sensitivity-specificity balance, achieving the best overall balanced accuracy and a tunable choice between maximal case-finding and minimal over-referral; and (iii) a transparent account of where small positive-class samples and dataset shift bound current methods.

## 2. Literature Review

### 2.1 Global and African burden of ROP

Modelled global estimates attribute a substantial share of preterm-associated visual impairment to ROP, with the largest projected increases occurring in regions undergoing rapid improvements in neonatal survival without commensurate scaling of screening and treatment(Blencowe et al., 2013). Gilbert (2008) framed this as successive "epidemics" of ROP, the first in high-income countries in the 1940s, the second in the 1970–80s due to increased survival of extremely preterm infants and now the 'third epidemic' in middle-income countries as neonatal care services expand(Gilbert, 2008). In these settings, broader gestational-age and birth-weight screening criteria are often required because larger, more mature babies may still develop severe disease, which both increases the screening workload and heightens the value of scalable, objective triage tools.

### 2.2 Plus disease and observer variability

Plus disease is defined by vascular dilation and tortuosity at the posterior pole and, together with zone and stage, determines treatment urgency under the revised international classification(Chiang et al., 2021). Because the judgement is qualitative and reference images are limited, expert agreement on the plus/pre-plus/normal distinction is imperfect(Campbell et al., 2021). This irreducible label noise places an upper bound on the achievable agreement of any automated system trained against expert labels and motivates evaluation metrics, such as quadratic-weighted kappa, that account for the ordered severity structure of the task.

### 2.3 Deep learning for plus-disease detection

Rao et al. (2018) demonstrated that a deep convolutional pipeline could diagnose plus disease at a level comparable to, and in some comparisons exceeding, that of human experts, establishing the feasibility of automated ROP grading in the Indian population(Rao et al., 2023). In parallel, Wang et al. (2018) developed DeepROP on a large Chinese screening dataset, reporting strong performance for identifying ROP and severe ROP(Wang et al., 2018). Reviews of artificial intelligence in ophthalmology have positioned ROP as one of the most mature application areas while cautioning that external validation, calibration, and prospective evaluation remain underdeveloped(Jafarizadeh et al., 2025).

### 2.4 Retinal vessel segmentation and architectures

Because plus disease is fundamentally a vascular phenotype, explicit vessel segmentation is a natural intermediate representation. The U-Net encoder-decoder and its nested variant U-Net++ are standard choices for biomedical segmentation(Zhou et al., 2018) and there are some implementations of its use in ROP(K.-W. Huang et al., 2023). With segmentation models, composite objectives combining cross-entropy, Dice, and centreline-aware (clDice) losses are commonly used to preserve thin, connected vessels. For classification, modern convolutional backbones such as EfficientNet(Tan & Le, 2020), DenseNet(G. Huang et al., 2018) and ResNet(Targ et al., 2016) provide a spectrum of capacity-efficiency trade-offs whose relative merits are dataset-dependent and therefore warrant empirical comparison on the target population.

### 2.5 Methodological pitfalls in medical AI

Several methodological hazards recur in clinical imaging AI and are especially consequential for small ROP datasets. First, splitting data at the image level rather than the patient level leaks information between training and test sets because multiple images of the same eye, and both eyes of the same infant, are highly correlated; this inflates apparent performance. Second, class imbalance here can cause naive models and accuracy-based selection to favour the majority class. Third, deep networks are frequently miscalibrated, so reported probabilities may not reflect true risk unless corrected, for example by temperature scaling(Guo et al., 2017). Fourth, focal and class-balanced losses and sensitivity-targeted thresholding are often needed to align optimisation with the clinical priority of minimising missed disease(Cui et al., 2019). The present study was designed explicitly to control each of these hazards.

# 3. Methodology

## 3.1 Study design and cohort

We conducted a retrospective comparative modelling study on a pilot ROP screening dataset from a Kenyan programme. After reconciling the clinical register with images available on disk, the analysable cohort comprised 121 preterm infants, 237 eyes, and 1,635 colour fundus images. Eye-level reference grades were No Plus (187 eyes, 78.9%), Pre-Plus (23 eyes, 9.7%), and Plus (27 eyes, 11.4%). All dataset counts were verified programmatically rather than assumed. The marked class imbalance and modest absolute number of Plus eyes are characteristic of single-centre pilot ROP datasets and are central to interpreting the results.

## 3.2 Reference standard and vessel annotations

Eye-level plus grades from the clinical register served as the classification reference standard. For the vasculature, two independent medical doctors did pixel level annotation by tracing out the retinal vasculature, demarcation zones and optic discs using CVAT. The principal-investigator graders(1 paediatric ophthalmologist and 1 vitro-retinal ophthalmologist) confirmed pixel-level vessel annotations in COCO format on separate, non-overlapping image sets. Because the two annotated sets contain different images (verified by content hashing; no byte-identical image appeared in both), they were treated as complementary supervision and pooled, rather than as repeated annotations of the same images; inter-grader agreement was therefore not applicable. Annotations also included optic-disc regions, enabling anatomically referenced (disc-centred) vascular zoning where present.

## 3.3 Leakage-resistant nested cross-validation

To prevent the patient-level leakage that inflates ROP-AI estimates, all partitioning was performed at the level of the infant (study identifier) using stratified group k-fold assignment, so that every image from both eyes and all visits of a given infant remained within a single fold. We used a nested design with five outer folds for unbiased evaluation and three inner folds for model selection, threshold optimisation, and calibration. Stratification used a patient-level severity surrogate (the maximum eye grade per infant) while training and evaluation retained eye-level targets. Per-fold class composition was recorded; rare Plus eyes were distributed across folds (4–10 Plus eyes per outer-test fold). Exact fold assignments were written to disk and reused across all experiments so that every architecture was compared on identical partitions.

### 3.4 Vessel segmentation

A U-Net++ segmenter with an EfficientNet-B0 encoder was trained to predict vessel probability maps from RGB images only; manual masks were used solely as supervision and never as model inputs. The objective combined binary cross-entropy, Dice, and centreline-aware (clDice) terms to preserve thin vessels. For each outer fold the segmenter was trained on the training infants, the binarisation threshold was selected on inner-validation data by maximising Dice, and predicted probability maps were generated for held-out infants. Mixed-precision training, channels-last memory layout, and multi-worker data loading were used for throughput.

### 3.5 Classification architectures compared

Eleven configurations were evaluated for eye-level Plus detection:

(1–6) six RGB backbones — EfficientNet-B0/B2/B3, DenseNet-121, ResNet-50, and ConvNeXt-Tiny — fine-tuned with class-weighted loss;

(7) an ordinal-severity model with cumulative-threshold outputs respecting the No Plus < Pre-Plus < Plus order;

(8) a gated-attention multiple-instance learning (MIL) model aggregating all images of an eye into an eye-level decision;

(9) a shared-encoder multi-task model jointly predicting vessel segmentation and Plus class;

(10) a temperature-calibrated EfficientNet-B2; and

(11) a two-stage "segment-then-classify" pipeline in which morphological features (vessel density, tortuosity, branching, calibre, fractal dimension) extracted from the predicted vessel maps were fed to a gradient-boosted binary classifier.

### Classification-model development and evaluation

The modelling pipeline evaluated eleven Plus-disease classification approaches using a patient-grouped cross-validation design. The clinical label was encoded as a three-class eye-level outcome: No Plus = 0, Pre-Plus = 1, and Plus = 2. Retinal images were loaded as three-channel RGB images, resized using longest-side scaling, padded to a square input size, normalised with ImageNet mean and standard deviation, and augmented during training using horizontal flipping, affine transformation, brightness/contrast perturbation, and hue-saturation-value adjustment. Elastic deformation was not used because of plus disease is dependent on vessel morphology which elasticity affects.

### RGB backbone classifiers: EfficientNet-B0/B2/B3, DenseNet-121, ResNet-50, and ConvNeXt-Tiny

The six RGB backbone classifiers were implemented using the timm model library through a common classifier wrapper. For each backbone, the original ImageNet classification layer was removed by creating the model with num_classes=0 and global average pooling. The resulting backbone acted as a feature extractor for each fundus image. A new task-specific classification head was added, consisting of dropout followed by a linear layer producing three logits for No Plus, Pre-Plus, and Plus disease. The evaluated backbones were EfficientNet-B0, EfficientNet-B2, EfficientNet-B3, DenseNet-121, ResNet-50, and ConvNeXt-Tiny. Each model received RGB images only; vessel masks were not concatenated as a fourth channel.

The RGB backbone classifiers were fine-tuned using class-weighted cross-entropy loss to account for the imbalance between No Plus, Pre-Plus, and Plus eyes. Class weights were calculated from the training fold only using inverse class frequency and normalised to have a mean of one. The models were trained with AdamW optimisation using discriminative learning rates: a lower learning rate for the pretrained backbone and a higher learning rate for the newly added classification head. The default configuration used a backbone learning rate of (1 ^{-5}), a head learning rate of (1 ^{-4}), and weight decay of (1 ^{-4}). Training used mixed precision with CUDA, gradient clipping with a maximum norm of 5.0, cosine learning-rate scheduling, and early stopping based on validation loss. The best validation-loss checkpoint was restored before generating test predictions.

For each RGB classifier, the model generated a three-class probability vector for every image. Probabilities from all images belonging to the same eye were aggregated into a single eye-level probability vector, with mean aggregation used for the principal analysis. The eye-level Plus probability was then compared with a fold-specific threshold selected on the validation set. The threshold-selection rule searched across 1,001 candidate thresholds from 0 to 1 and chose the threshold that maximised specificity while meeting the target Plus sensitivity, set by default at 90%. If no threshold met the target sensitivity, the threshold with the highest available sensitivity was retained and flagged as not meeting the target. A separate threshold was also selected for referable disease, defined as Pre-Plus or Plus versus No Plus, using (P() + P()).

## Ordinal-severity model

The ordinal-severity model was implemented as a cumulative-threshold classifier to reflect the ordered clinical relationship No Plus < Pre-Plus < Plus. The model used an RGB backbone, typically EfficientNet-B2, as the feature extractor. Instead of producing three independent class logits, the classification head produced two cumulative logits corresponding to (P(Y )) and (P(Y )). The three clinical labels were converted into cumulative binary targets: No Plus was encoded as [0,0], Pre-Plus as [1,0], and Plus as [1,1]. The model was trained using binary cross-entropy with logits over these two cumulative outputs. At inference, sigmoid activation was applied to the two logits, and monotonicity was enforced so that (P(Y )) could not exceed (P(Y )). The cumulative outputs were then converted back into a three-class probability distribution before eye-level aggregation and sensitivity-targeted thresholding.

## Gated-attention multiple-instance learning model

The multiple-instance learning model was implemented at the eye level rather than the image level. Each eye was treated as a variable-sized bag containing all retinal images available for that eye. Each image in the bag was passed through a shared CNN encoder(EfficientNet-B2) to produce an image embedding. These embeddings were then passed through a gated-attention pooling module. The attention module learned an importance weight for each image using parallel tanh and sigmoid transformations followed by a learned scoring layer and softmax normalisation. The weighted image embeddings were summed to create a single eye-level embedding, which was passed through dropout and a linear classification head to produce three logits for No Plus, Pre-Plus, and Plus disease. Because eyes had different numbers of images, the model was trained using one eye-bag per batch. The MIL model therefore produced an eye-level prediction directly, without requiring post-hoc averaging of image-level probabilities. Attention weights were retained as an interpretability output, including the maximum attention value per eye.

## Shared-encoder multitask model

The shared-encoder multitask model was implemented to jointly learn retinal-vessel segmentation and Plus-disease classification. The architecture used a U-Net model from segmentation-models-pytorch with an EfficientNet encoder. The RGB image was passed through a shared encoder, after which two task-specific outputs were generated. The segmentation branch used the decoder and segmentation head to produce a one-channel vessel logit map. The classification branch used the deepest encoder feature map, applied adaptive average pooling, flattened the result, and passed it through a classification head consisting of dropout, a 256-unit fully connected layer, ReLU activation, a second dropout layer, and a final three-class linear layer. Manual vessel masks were used only as segmentation targets and were not used as model inputs. The total training loss combined cross-entropy classification loss with a weighted segmentation loss. In the executed configuration, the segmentation loss was weighted using a segmentation weight, with 0.25 used as the default setting.

During multitask training, each batch contained RGB images, eye-level disease labels, and vessel masks where available. The classification loss was calculated from the three-class disease logits. The segmentation loss was calculated from the predicted vessel logits and the corresponding vessel mask after resizing the mask to the segmentation-output resolution when needed. The segmentation component combined binary cross-entropy with Dice-based overlap loss and was multiplied by the specified segmentation weight before being added to the classification loss. After training, the model's classification branch was used to generate image-level disease probabilities on the validation and test sets. These probabilities were aggregated to the eye level using the same mean aggregation approach applied to the RGB backbone classifiers. Fold-specific Plus and referable-disease thresholds were selected on validation predictions and applied unchanged to the held-out test eyes.

## Temperature-calibrated EfficientNet-B2

The temperature-calibrated model used the standard EfficientNet-B2 RGB classifier as the base model and then applied post-hoc probability calibration. The EfficientNet-B2 classifier was trained using the same RGB input pipeline, class-weighted cross-entropy loss, patient-grouped folds, AdamW optimiser, and early-stopping framework used for the other RGB backbones. After model training, raw image-level logits were collected from the validation set and averaged to the eye level. A single scalar temperature parameter was then fitted on the validation eye-level logits using the true validation labels. The fitted temperature was applied to the outer-test eye-level logits before softmax transformation, producing calibrated class probabilities. Calibration performance was assessed using Brier score, expected calibration error, and negative log-likelihood. The calibration procedure therefore assessed whether the EfficientNet-B2 probability outputs could be made more reliable without changing the underlying model architecture or retraining the classifier.

### Vessel segmentation model

The vessel segmentation component was implemented as a standalone RGB-to-vessel-mask model using segmentation-models-pytorch. The primary configuration used U-Net++ with an EfficientNet-B0 encoder. The segmenter received RGB fundus images and produced a one-channel vessel logit map. Manual vessel annotations were used as segmentation supervision only. The segmentation loss combined binary cross-entropy, soft Dice loss, and centreline Dice loss, with default weights of 0.4, 0.4, and 0.2 respectively. For each patient-grouped outer fold, the segmenter was trained on annotated images from the training patients. A binarisation threshold was selected on the validation images by maximising mean Dice coefficient across candidate thresholds from 0.20 to 0.80. The selected threshold was then applied to the held-out test images. The code saved out-of-fold vessel-probability maps, per-image segmentation metrics, and visual overlay panels showing true-positive vessel pixels in green, false negatives in blue, and false positives in red.

### Two-stage segment-then-classify pipeline

The segment-then-classify pipeline was implemented as a deployable two-stage approach for Plus versus non-Plus classification. In the first stage, the vessel segmentation model generated out-of-fold vessel-probability maps from RGB fundus images. These maps were binarised using the corresponding validation-selected segmentation threshold. In the second stage, morphological vessel features were extracted from the predicted binary vessel masks. The code calculated features including vessel area fraction, vessel density, total skeleton length, normalised skeleton length, vessel-width summaries, endpoint count, branchpoint count, endpoint density, branch density, branch-to-endpoint ratio, fractal dimension, number of vessel segments, arc-to-chord tortuosity, tortuosity variability, curvature-related tortuosity, vessel-segment length summaries, and regional vessel densities. The vessel skeleton was generated from the binary mask; branchpoints were identified and removed before individual vessel segments were labelled for tortuosity calculation. Vessel width was estimated using the Euclidean distance transform along the skeleton.

After feature extraction, the segment-then-classify pipeline trained a binary gradient-boosting classifier to distinguish Plus disease from No Plus or Pre-Plus. Feature scaling was performed using a StandardScaler fitted on the training fold only. The classifier was a scikit-learn GradientBoostingClassifier trained on image-level vessel-morphology features with the binary target (Y=1) for Plus and (Y=0) for No Plus or Pre-Plus. The classifier produced an image-level Plus probability, and image probabilities were then aggregated to the eye level using mean aggregation by default. A Plus threshold was selected on the validation eyes using the same sensitivity-targeted thresholding rule as the CNN models and was applied unchanged to the held-out test eyes. The code was designed to use predicted out-of-fold vessel masks when available. If predicted masks were not found, it could fall back to manual masks, but this was explicitly labelled as an upper-bound, non-deployable analysis rather than a true inference setting.

### 3.6 Eye-level aggregation, thresholds, calibration, and metrics

Image-level probabilities were aggregated to the eye (mean by default; attention pooling for the MIL model). Following the clinical imperative to minimise missed disease, decision thresholds targeted 90% Plus sensitivity, were selected on inner-validation data, and were then applied unchanged to the outer-test fold; a separate referable-disease threshold (Pre-Plus or Plus versus No Plus) was selected independently. Probability calibration used temperature scaling fitted on validation logits. The primary endpoint was eye-level Plus versus non-Plus, reported as sensitivity, specificity, macro AUPRC, and quadratic-weighted kappa. Uncertainty was quantified by patient-grouped bootstrap resampling of infants (2,000 replicates), and the two leading deployable models were compared with a paired patient-bootstrap difference test. No threshold, backbone, or hyper-parameter was ever selected on outer-test data.

### 3.7 Combined reading: ensemble and two-step analysis

To test whether the two model families are complementary, we combined a representative vision classifier (the RGB baseline) with a representative segmentation-coupled model (the shared-encoder multi-task model) on the eyes for which both produced predictions. Three combination strategies were evaluated. A probability ensemble averaged the two models' Plus probabilities. A two-step OR reading ("sensitive screen") classified an eye as Plus if either model did so. A two-step AND reading ("specific confirmation") required both models to agree. To avoid any optimism, decision thresholds for each arm were chosen by a leave-one-outer-fold-out procedure — the Youden-optimal threshold was selected on the other folds and applied to the held-out fold, then predictions were pooled — so no eye was ever thresholded using its own fold. Sensitivity, specificity, and balanced accuracy were reported with patient-grouped bootstrap confidence intervals (2,000 replicates resampling infants). The two-stage segment-then-classify pipeline was examined analogously as a robustness check, although its smaller shared-eye sample limited precision.

### 3.8 Implementation

Models were implemented in PyTorch with the segmentation-models-pytorch and timm libraries and trained on a single NVIDIA RTX 4060 Ti (16 GB) GPU with mixed precision. Reproducibility controls fixed random seeds and recorded the full software environment. The complete pipeline, including the data audit, fold assignments, training and evaluation scripts, and reporting, is available on request.

## 4. Results

### 4.1 Cohort and data audit

The analysed ROP dataset comprised 121 preterm infants contributing 237 eyes, with 116 infants having bilateral eye records and five infants contributing only one eye. Laterality was well balanced, with 117 left eyes and 120 right eyes. The dataset contained a reported total of 1,883 retinal images, with a median of 7 images per eye and a mean of 7.9 images per eye, ranging from 2 to 22 images.

The cohort had a mean gestational age at birth of 29.5 weeks, with a median of 30.0 weeks and a range from 24.0 to 34.0 weeks. The mean postmenstrual age at examination was 38.6 weeks, with a median of 38.0 weeks and a range from 28.7 to 53.0 weeks. At the eye level, most examinations were classified as No Plus disease, accounting for 187 of 237 eyes (78.9%). Pre-Plus disease was present in 23 eyes (9.7%), while Plus disease was present in 27 eyes (11.4%), giving a combined referable vascular abnormality group of 50 eyes (21.1%).

At the infant level, using the most severe eye grade per infant, 90 infants (74.4%) had No Plus disease, 12 (9.9%) had Pre-Plus disease, and 19 (15.7%) had Plus disease.

*Table 1. Descriptive characteristics of the ROP dataset*

| **Characteristic** | | **Summary(n=1,883)** |
|---|---|---|
| **Dataset structure** | Infants | 121 |
| | Eyes | 237 |
| | Bilateral eye records | 116/121 infants, 95.9% |
| | Images per eye, mean ± SD | 7.9 ± 2.5 |
| **Laterality** | Left eye | 117/237, 49.4% |
| | Right eye | 120/237, 50.6% |
| **Gestational age at birth** | Median (IQR), weeks | 30.0 (24.0–34.0) |
| | <28 weeks | 27/237 eyes, 11.4% |
| | 28 to <32 weeks | 174/237 eyes, 73.4% |
| | ≥32 weeks | 36/237 eyes, 15.2% |
| | | |
| **Postmenstrual age at examination** | Median (Range), weeks | 38.0 (28.7–53.0) |
| **Plus-disease classification, eye-level** | No Plus | 187/237, 78.9% |
| | Pre-Plus | 23/237, 9.7% |
| | Plus | 27/237, 11.4% |
| | Referable vascular abnormality, Pre-Plus or Plus | 50/237, 21.1% |
| **Worst-eye Plus-disease classification, infant-level** | No Plus | 90/121, 74.4% |
| | Pre-Plus | 12/121, 9.9% |
| | Plus | 19/121, 15.7% |
| | Referable vascular abnormality, Pre-Plus or Plus | 31/121, 25.6% |

## 4.2 Vessel segmentation

The U-Net++ vessel segmenter, trained on RGB images alone, produced coherent vascular maps on Kenyan fundus images. Pooled across 1,029 held-out images from the five outer folds, it achieved a Dice coefficient of 0.533 and an intersection-over-union of 0.368, with high pixel specificity (0.979) and moderate pixel sensitivity (0.623) and precision (0.481). Per-fold Dice was stable, ranging from 0.513 to 0.549, indicating consistent behaviour across patient partitions rather than reliance on a single favourable split.

*Table 2. Vessel segmentation performance (U-Net++ / EfficientNet-B0) on held-out images. Thresholds were chosen on inner-validation by maximising Dice.*

| Outer fold | Images | Threshold | Dice | IoU | Pixel Sens. | Pixel Spec. |
|---|---|---|---|---|---|---|
| 0 | 197 | 0.45 | 0.528 | 0.365 | 0.620 | 0.980 |
| 1 | 228 | 0.30 | 0.549 | 0.381 | 0.647 | 0.976 |
| 2 | 206 | 0.55 | 0.541 | 0.373 | 0.649 | 0.979 |
| 3 | 207 | 0.60 | 0.532 | 0.366 | 0.621 | 0.979 |
| 4 | 191 | 0.60 | 0.513 | 0.352 | 0.569 | 0.983 |
| **Pooled** | **1029** | — | **0.533** | **0.368** | **0.623** | **0.979** |

Qualitative review of the overlay panels (Figure 1) showed that the model reliably recovered the disc-centred radiating arcade and major branches, with errors concentrated at thin distal vessels (false negatives, blue) and at vessel borders (false positives, red). The predicted probability maps closely tracked the manual annotations in well-exposed images and degraded gracefully under haze or peripheral illumination falloff, where faint distal vessels were missed rather than hallucinated.

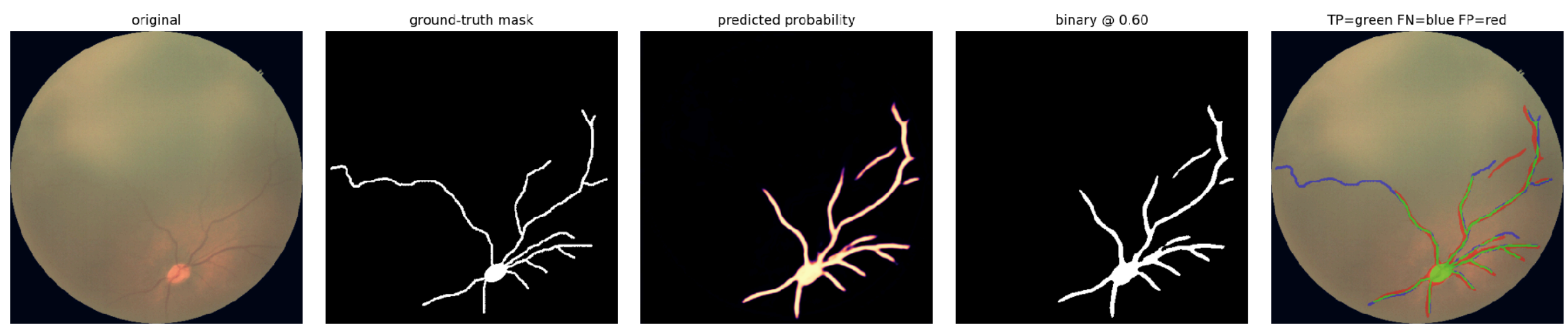

*Figure 1a. Representative held-out vessel segmentation (left eye). Panels: original image, manual annotation, predicted probability, binarised prediction, and an error overlay (true positives green, false negatives blue, false positives red).*

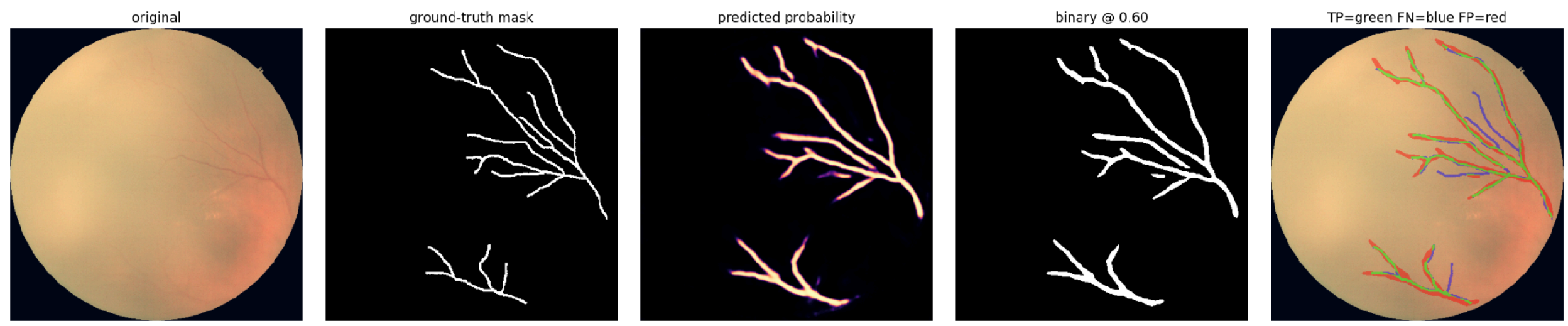

*Figure 1b. Representative held-out vessel segmentation (right eye), showing accurate recovery of the disc and major arcade with distal-vessel false negatives.*

## 4.3 Plus-disease classification: model comparison

Across the eleven configurations, eye-level Plus detection exhibited a pronounced sensitivity-specificity trade-off (Table 3). The shared-encoder multi-task model attained the highest mean macro AUPRC (0.685), the metric least distorted by the class imbalance, combining high sensitivity (0.900) with the best specificity among the high-sensitivity models. The ConvNeXt-Tiny backbone achieved the next-highest AUPRC (0.630) and the strongest discrimination of ordered severity (quadratic-weighted kappa 0.327) and the highest specificity overall (0.684), but at lower sensitivity (0.717). At the opposite extreme, the temperature-calibrated EfficientNet-B2 and the MIL model reached very high or perfect sensitivity (1.000 and 0.950) but collapsed in specificity (0.178 and 0.102), indicating near-universal positive prediction. Standard backbones (EfficientNet-B2/B3, ResNet-50, DenseNet-121) clustered at high sensitivity (0.78–0.93) with low-to-moderate specificity.

*Table 3. Eye-level Plus-detection performance, mean across five patient-grouped outer folds, ranked by macro AUPRC.*

| Model | Sens. | Spec. | Macro AUPRC | Wt. kappa |
|---|---|---|---|---|
| Multi-task (Eff-B2) | 0.900 | 0.478 | 0.685 | — |
| ConvNeXt-Tiny | 0.717 | 0.684 | 0.630 | 0.327 |
| EfficientNet-B3 | 0.900 | 0.224 | 0.567 | 0.204 |
| Ordinal (Eff-B2) | 0.878 | 0.449 | 0.538 | — |
| EfficientNet-B2 | 0.892 | 0.353 | 0.521 | 0.268 |
| EfficientNet-B0 | 0.856 | 0.380 | 0.511 | 0.212 |
| ResNet-50 | 0.933 | 0.205 | 0.483 | 0.000 |
| DenseNet-121 | 0.783 | 0.280 | 0.444 | 0.063 |
| MIL (Eff-B2) | 0.950 | 0.102 | 0.305 | — |
| Seg-then-classify (GBM) | 0.733 | 0.487 | 0.272 | — |
| Calibrated (Eff-B2) | 1.000 | 0.178 | — | — |

*Dashes indicate metrics not defined for that configuration. Sens. = sensitivity; Spec. = specificity; Wt. kappa = quadratic-weighted kappa.*

## 4.4 Uncertainty and paired comparison of deployable models

For the five configurations producing per-eye predictions, patient-grouped bootstrap resampling (2,000 replicates) quantified uncertainty (Table 4). The corrected RGB baseline (ConvNeXt-Tiny) achieved 0.692 Plus sensitivity (95% CI 0.455–0.900) and 0.667 specificity (0.589–0.745). The two-stage segment-then-classify pipeline reached 0.750 sensitivity (0.500–0.953) at 0.485 specificity. The ordinal and multi-task models, at their selected thresholds, were highly specific (0.990 and 0.995) but insensitive (0.077 and 0.385), reflecting conservative operating points. A paired patient-bootstrap comparison of the segment-then-classify pipeline against the RGB baseline found a sensitivity difference of +0.188 whose 95% confidence interval (−0.174 to 0.545) included zero — the two were not statistically distinguishable on sensitivity — while the specificity difference of −0.187 (−0.296 to −0.077) excluded zero, indicating the two-stage pipeline was significantly less specific on this cohort.

*Table 4. Patient-grouped bootstrap estimates (2,000 replicates) for the deployable models at their selected operating points*

| Model | Plus sensitivity (95% CI) | Plus specificity (95% CI) |
|---|---|---|
| RGB baseline (ConvNeXt) | 0.692 (0.455–0.900) | 0.667 (0.589–0.745) |
| Segment-then-classify | 0.750 (0.500–0.953) | 0.485 (0.397–0.569) |
| Multi-task (Eff-B2) | 0.385 (0.182–0.594) | 0.995 (0.985–1.000) |
| MIL (Eff-B2) | 0.154 (0.000–0.333) | 0.948 (0.914–0.977) |
| Ordinal (Eff-B2) | 0.077 (0.000–0.182) | 0.990 (0.976–1.000) |

*Operating points differ by design, so sensitivity and specificity must be read jointly.*

### 4.5 Calibration

Temperature scaling produced modest improvements in probability calibration in four of five folds, reducing the Brier score (for example from 0.532 to 0.517 and from 0.452 to 0.425) and the negative log-likelihood. In one fold the validation-fitted temperature degenerated toward zero and worsened test calibration, a known instability when validation logits are few and the positive class is small. Calibration therefore improved reliability on average but was not uniformly beneficial, underscoring that calibration must itself be validated per deployment fold rather than assumed.

### 4.6 Combined reading: complementarity of classification and segmentation

The two model families proved complementary, and combining them allowed the operating point to be set deliberately (Table 5). On the 236 eyes for which both the vision classifier and the segmentation-coupled multi-task model produced predictions, the vision classifier alone reached a sensitivity of 0.615 and specificity of 0.781, while the segmentation-coupled model alone was substantially more specific (0.914) at comparable sensitivity (0.692) — consistent with the broader pattern that classifiers over-refer whereas vessel-aware models are conservative.

A probability ensemble of the two achieved the best single balance, with a balanced accuracy of 0.803, exceeding the vision classifier alone (0.698). The two-step readings then traded sensitivity against specificity as intended: the OR “sensitive screen” reached the highest sensitivity of any configuration (0.769, 95% CI 0.550–0.950), maximising case-finding, whereas the AND “specific confirmation” reached the highest specificity (0.957, 0.929–0.981), minimising over-referral. Thresholds were selected by leave-one-fold-out so these operating points are free of test-set tuning.

*Table 5. Combined reading of a vision classifier and a segmentation-coupled model on shared eyes (leave-one-fold-out Youden thresholds; patient-grouped bootstrap, 2,000 replicates).*

| Reading strategy | Sensitivity (95% CI) | Specificity (95% CI) | Bal. acc. |
|---|---|---|---|
| Vision classifier alone | 0.615 (0.350–0.833) | 0.781 (0.716–0.846) | 0.698 |
| Segmentation model alone | 0.692 (0.469–0.886) | 0.914 (0.876–0.948) | 0.803 |
| Probability ensemble | 0.692 (0.462–0.885) | 0.914 (0.873–0.951) | 0.803 |
| **Two-step OR (sensitive screen)** | **0.769 (0.550–0.950)** | **0.738 (0.670–0.802)** | **0.754** |
| Two-step AND (specific confirm) | 0.538 (0.300–0.769) | 0.957 (0.929–0.981) | 0.748 |

*The ensemble gives the best single balance; the OR and AND readings deliberately maximise sensitivity and specificity respectively. Bal. acc. = balanced accuracy (mean of sensitivity and specificity).*

A robustness check substituting the two-stage segment-then-classify pipeline for the multi-task model showed the same qualitative pattern — the AND reading again produced the highest specificity (0.941) — but rested on only 38 shared eyes, so its confidence intervals were wide and it is reported as directional rather than confirmatory. Together these results support the study's central hypothesis: classification and segmentation contribute sensitivity and specificity respectively, and a combined reading converts that complementarity into a controllable screening operating point.

## 5. Discussion

This study's central finding is that image classification and vessel segmentation are complementary readers for ROP Plus screening, not competing alternatives. Across a leakage-resistant, patient-grouped benchmark on a sub-Saharan African preterm cohort, RGB classifiers were the sensitive family catching most Plus eyes but over-referring. Segmentation-coupled models were the specific family, exemplified by the multi-task model's 0.914 specificity and the vessel segmenter's 0.979 pixel specificity. Combining the two converted this complementarity into a controllable operating point: a probability ensemble achieved the best single balance (balanced accuracy 0.803), an OR “sensitive screen” maximised case-finding sensitivity (0.769), and an AND “specific confirmation” maximised specificity (0.957). This is the practical headline for programmes: a combined classifier-plus-segmentation reader lets a service choose where on the sensitivity-specificity curve it wishes to operate, rather than being locked to a single model's trade-off.

Two supporting findings reinforce this message. First, automated vessel segmentation transfers to Kenyan fundus images: a U-Net++ model trained from RGB alone recovered the posterior arcade with a pooled Dice of 0.533 and stable per-fold behaviour, capturing precisely the vascular structures that define plus disease and providing the high-specificity signal the combined reader depends on. Second, among stand-alone classifiers the segmentation-aware multi-task model achieved the best macro AUPRC (0.685), indicating that coupling segmentation to classification helps even before any explicit ensembling, while ConvNeXt-Tiny offered the best ordered-severity agreement. The recurrent sensitivity-specificity trade-off seen across all single models is exactly the problem the combined reading is designed to manage.

These results should be read against the established literature rather than in isolation. Systems such as i-ROP DL report higher discrimination for plus disease(Redd et al., 2018) , and DeepROP likewise reports strong performance on a large cohort (Wang et al., 2018). The gap is unsurprising and instructive: those systems were developed on substantially larger datasets with hundreds of treatment-requiring cases, whereas the present cohort contains only 27 Plus eyes, and they were evaluated largely within-distribution. The KIDROP study for example used 11 years worth of data compared to our study with less than an year's worth of data collection(Rao et al., 2023). Our more modest absolute numbers are therefore best interpreted not as a ceiling on what is achievable in African settings, but as a realistic baseline established under honest, leakage-free evaluation on a small pilot cohort; the conditions most programmes will actually face at inception.

Clinically, the asymmetry of error costs is decisive. A missed Plus eye risks irreversible blindness, whereas a false positive incurs an additional specialist review. This justifies the sensitivity-prioritised thresholds used here, but the resulting false-positive volume (70 over-referrals pooled) clarifies that such models are tools for triage and workload reduction within a human-in-the-loop pathway, not autonomous diagnosis. The calibrated model's collapse to near-universal positivity is the limiting case of this logic and would, in practice, refer almost everyone. The multi-task and ConvNeXt configurations are more promising precisely because they retain usable specificity while remaining sensitive, which is what determines whether automated triage actually reduces specialist burden.

Several limitations bound these conclusions. The cohort is a single-centre pilot with few positive cases, so confidence intervals are wide and per-fold estimates are volatile; the wide bootstrap intervals (for example sensitivity 0.455–0.900 for the baseline) are an honest reflection of this. Eye-level grades are subject to the inter-observer variability intrinsic to plus disease, imposing a label-noise ceiling on attainable agreement. There is no external or prospective validation, and image quality was not formally gated, so some errors likely reflect acquisition rather than model capacity. Strengths counterbalance these: testing of multiple model architectures, strict patient-grouped nested cross-validation and sensitivity-targeted thresholds fixed without test-set access. This is the first Kenyan ROP study done comparing various AI architectures for screening.

## 6. Recommendations

Translating these findings into clinically useful tools for African ROP programmes implies several priorities. First and foremost, build screening systems as combined classifier-plus-segmentation readers rather than single models: pair a sensitive image classifier, which finds candidate disease, with a specific vessel-segmentation model, which suppresses false referrals, and let the programme select the operating point — an OR "sensitive screen" where missing disease is least acceptable, an AND "specific confirmation" where referral capacity is scarce, or a probability ensemble for the best overall balance.

Second, assemble larger, multi-site African datasets with harmonised grading, deliberately enriching for treatment-requiring disease; the dominant constraint here is positive-case scarcity, not model architecture, and federated or data-sharing collaborations across centres would address it without centralising sensitive images.

Third, retain explicit vessel segmentation as a core component, both because it supplies the specificity the combined reader depends on and because an auditable vessel map supports clinician trust. Fourth, embed the combined reader in a human-in-the-loop pathway and report the expected over-referral burden alongside sensitivity, so programmes can size review capacity realistically. Fifth, treat calibration as a first-class, per-deployment concern, validating it on local data rather than assuming transferability, given the instability observed in one fold. Sixth, gate on image quality at acquisition, since several confident errors co-occurred with poor exposure.

Finally, validate prospectively with patient-grouped uncertainty and pre-registered operating points, and sustain transparent, leakage-resistant evaluation with open code so that comparisons across African cohorts remain meaningful as datasets grow.

## 7. Project Next Steps

The authors are in the process of increasing the dataset quantitatively and qualitatively. The qualitative increase will involve the incorporation of ROP risk factors like gestational age, twinning, oxygen exposure etc to the dataset to evaluate contrastive models and multimodal models. This will also include a higher degree of annotation of the retinal images simultaneously by multiple readers.

## 8. Conclusion

Image classification and vessel segmentation are complementary, not competing, approaches to ROP Plus-disease screening: on a real Kenyan preterm cohort, classifiers supplied sensitivity while segmentation supplied specificity, and combining them in a probability ensemble or a two-step reading produced a controllable operating point — the best overall balance from the ensemble, the highest sensitivity from an OR "sensitive screen", and the highest specificity from an AND "specific confirmation". Vessel segmentation itself transferred successfully to African fundus images, and segmentation-aware classification gave the best stand-alone precision-recall balance among eleven configurations evaluated under strict, leakage-resistant cross-validation. Performance remains bounded by a small positive-class sample and dataset shift, so the path to clinically trustworthy systems runs through larger, multi-site African datasets and prospective validation. The practical recommendation, however, is clear: ROP screening tools should be built as combined classifier-plus-segmentation readers, deployed with a sensitive first pass and a specific confirmation within a human-in-the-loop pathway. Established under transparent evaluation, the results here offer both a realistic baseline and a concrete design direction for that work.

## 9. Disclosures

Funding for the original image acquisition and annotation was provided by the College of Ophthalmology of Eastern Central and Southern Africa while the AI technical support and compute costs were provided by the Center for Health AI Research Innovation and Implementation.